\documentclass{article}

\usepackage[dblblindworkshop, final]{neurips_2025}

\usepackage[utf8]{inputenc} 
\usepackage[T1]{fontenc}    
\usepackage{hyperref}       
\usepackage{url}            
\usepackage{booktabs}       
\usepackage{amsfonts}       
\usepackage{nicefrac}       
\usepackage{microtype}      
\usepackage{xcolor}         

\usepackage{microtype}
\usepackage{hyperref}
\usepackage{url}
\usepackage{booktabs}

\usepackage{lineno}

\definecolor{darkblue}{rgb}{0, 0, 0.5}
\definecolor{illiniorange}{HTML}{FF5F05}
\hypersetup{colorlinks=true, citecolor=illiniorange, linkcolor=illiniorange, urlcolor=illiniorange}

\usepackage{graphicx}
\usepackage{wrapfig}
\usepackage{tabularx}
\usepackage{colortbl}
\usepackage{xspace}
\usepackage{xcolor}
\usepackage{multirow}
\usepackage{multicol}
\usepackage{amsmath}
\usepackage{enumitem}
\definecolor{tableheader}{HTML}{EFEFEF}
\newcommand{\ours}{\textsc{AURA}\xspace}
\newcommand{\ie}{\textit{i.e.}}
\newcommand{\eg}{\textit{e.g.}}
\renewcommand{\sectionautorefname}{Section}

\usepackage{listings}
\newcommand{\benchone}{TravelPlanner\xspace}
\newcommand{\benchtwo}{$\tau$-Bench-Airline\xspace}
\newcommand{\benchthree}{$\tau$-Bench-Retail\xspace}
\newcommand{\taubench}{$\tau$-Bench\xspace}

\newcommand{\fouro}{\texttt{gpt-4o}\xspace}
\newcommand{\fouromini}{\texttt{gpt-4o-mini}\xspace}
\newcommand{\sonnet}{\texttt{sonnet-3.5}\xspace}
\newcommand{\gemini}{\texttt{gemini-1.5-fsh.}\xspace}
\newcommand{\geminifull}{\texttt{gemini-1.5-flash}\xspace}
\newcommand{\threeturbo}{\texttt{gpt-3.5-turbo}\xspace}
\newcommand{\mlarge}{\texttt{mistral-large}\xspace}
\newcommand{\mixtral}{\texttt{mixtral-8x7B}\xspace}
\newcommand{\llama}{\texttt{llama-3.3-70B}\xspace}
\newcommand{\qwen}{\texttt{qwen2.5-72B}\xspace}

\newcommand{\rankone}{100}
\newcommand{\ranktwo}{60}
\newcommand{\rankthree}{40}
\newcommand{\rankfour}{25}
\newcommand{\rankfive}{10}
\newcommand{\colorprop}{cyan}
\definecolor{MyDarkGreen}{rgb}{0.0, 0.7, 0.0}
\newcommand{\coloros}{MyDarkGreen}

\NewDocumentCommand{\tk}
{ mO{} }{\textcolor{blue}{\textsuperscript{\textit{TK}}\textsf{\small[#1]}}}
\NewDocumentCommand{\dilek}
{ mO{} }{\textcolor{teal}{\textsuperscript{\textit{DHT}}\textsf{\small[#1]}}}
\NewDocumentCommand{\emre}
{ mO{} }{\textcolor{red}{\textsuperscript{\textit{Emre}}\textsf{\small[#1]}}}

\title{\ours: A Diagnostic Framework for Tracking\\ User Satisfaction of Interactive  Planning Agents}

\author{%
  Takyoung Kim\thanks{Equal Contribution.} \qquad Janvijay Singh$^*$ \qquad Shuhaib Mehri$^*$ \\ \textbf{Emre Can Acikgoz} \quad \textbf{Sagnik Mukherjee} \quad \textbf{Nimet Beyza Bozdag} \quad \textbf{Sumuk Shashidhar} \\
  \textbf{Gokhan Tur} \qquad \textbf{Dilek Hakkani-Tür}
  \\
  University of Illinois Urbana-Champaign\\
  \texttt{\{tk30, jvsingh2, mehri2, gokhan, dilek\}@illinois.edu} \\
}

\begin{document}

\maketitle

\begin{abstract}
The growing capabilities of large language models (LLMs) in instruction-following and context-understanding lead to the era of agents with numerous applications. Among these, task planning agents have become especially prominent in realistic scenarios involving complex internal pipelines, such as context understanding, tool management, and response generation. However, existing benchmarks predominantly evaluate agent performance based on task completion as a proxy for overall effectiveness. We hypothesize that merely improving task completion is misaligned with maximizing user satisfaction, as users interact with the entire agentic process and not only the end result. To address this gap, we propose \textbf{\ours}, an \underline{A}gent-\underline{U}ser inte\underline{R}action \underline{A}ssessment framework that conceptualizes the behavioral stages of interactive task planning agents. \ours offers a comprehensive assessment of agent through a set of atomic LLM evaluation criteria, allowing researchers and practitioners to diagnose specific strengths and weaknesses within the agent's decision-making pipeline. Our analyses show that agents excel in different behavioral stages, with user satisfaction shaped by both outcomes and intermediate behaviors. We also highlight future directions, including systems that leverage multiple agents and the limitations of user simulators in task planning.
\end{abstract}
\section{Introduction}

Large language models (LLMs) are increasingly deployed in real-world applications, primarily due to their ability to understand complex goals and devise structured sequences of action: a process known as ``planning''~\cite{Wang_2024}. To assess and refine planning skills, researchers have introduced diverse benchmarks in web-based~\cite{NEURIPS2022_82ad13ec, xie2024openagents}, mobile~\citep{deng-etal-2024-mobile}, embodied~\citep{choi2024lotabench}, and automated testbeds~\citep{Park2023GenerativeAgents}. 

One of the most impactful planning applications is \textit{task planning}, where agents generate and execute domain-specific plans (\eg, itineraries) to help users achieve goals while adhering to contextual constraints~\citep{budzianowski-etal-2018-multiwoz, Rastogi_Zang_Sunkara_Gupta_Khaitan_2020, pmlr-v235-xie24j}. While agents must accurately interpret user requests, leverage predefined tools, and engage in personalized dialogues, many benchmarks focus solely on final task completion, overlooking the agents' intermediate behaviors and planning steps. 
Given that \textbf{overall user satisfaction is shaped by multiple factors} (\eg, efficiency and effectiveness) \textbf{throughout prolonged interactions}~\citep{ASHFAQ2020101473}, a narrow focus on outcome-based metrics can misrepresent an agent’s true effectiveness, especially in complex task-oriented applications~\citep{cao2025revisitingllmevaluationmechanism, xia2025evaluationdrivendevelopmentllmagents, acikgoz-etal-2025-td}. 

In addition, existing benchmarks across different domains often introduce \textit{bespoke} evaluation frameworks and metrics, making it possible for the same agent to be evaluated using entirely different criteria. This fragmentation introduces inconsistencies in how agent behavior is interpreted and makes it difficult to draw generalizable conclusions about agent capabilities. Without a domain-agnostic evaluation framework, it becomes challenging to meaningfully compare systems, track progress over time, and identify fundamental limitations in agent design. 

To address these limitations, we propose \textbf{\ours}, an \underline{A}gent-\underline{U}ser inte\underline{R}action \underline{A}ssessment framework. To ensure generalizability across sequential agentic scenarios, \ours is designed based on the partially observed Markov Decision Process (POMDP). This formulation reflects the reality that key aspects of interaction, such as user intention and satisfaction, are often hidden or only partially observable. Moreover, it aligns with the observation that contemporary agent behaviors typically unfold in a sequential, decision-making context consistent with the POMDP paradigm~\citep{10610981, 10.5555/3692070.3692783, pmlr-v235-xie24j, yao2025taubench}. Lastly, \ours defines domain-agnostic LLM evaluation criteria that can be easily instantiated through a set of atomic and easily measurable metrics. As illustrated in \autoref{fig:overview}, \ours provides the following key advantages: 
\begin{wrapfigure}[22]{r}{0.5\textwidth}
    \vspace{5mm}
    \centering
    \includegraphics[width=0.5\textwidth]{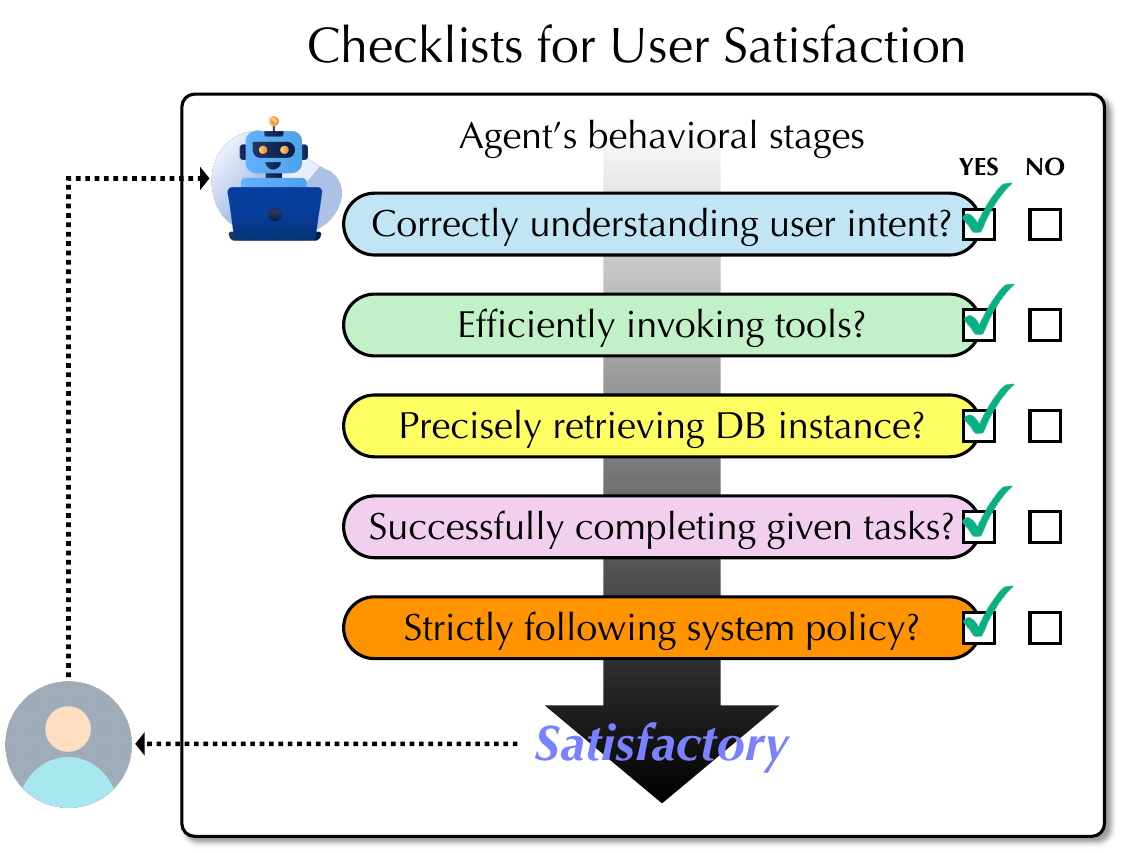}
    \caption{\ours provides unified, atomic, and domain-agnostic criteria for assessing user satisfaction of interactive planning agents, extending beyond conventional evaluation protocols that focus solely on task completion.}
    \label{fig:overview}
\end{wrapfigure}


\begin{enumerate}[leftmargin=*]
    \item \textbf{Generalizable Evaluation}: \ours establishes a domain-agnostic protocol for interactive task planning, enabling a consistent comparison across diverse benchmarks under a shared set of principles, which has been discussed for decades~\citep{walker-etal-1997-paradise} yet remains an open issue.

    \item	\textbf{Multi-Axis Diagnosis}: By aligning with the POMDP framework, \ours supports evaluation across multiple behavioral stages of the interaction pipeline, encompassing both intermediate decisions and final outcomes. This allows for a fine-grained diagnostic analysis of how different agents behave and impact performance.

    \item \textbf{Cross-Benchmark Comparisons}: \ours enables systematic comparisons across heterogeneous tasks and environments, thereby uncovering planning strategies and revealing performance trade-offs that may not be observable within the scope of individual benchmarks.
\end{enumerate}

Through extensive experiments, we demonstrate that different models exhibit distinct strengths and weaknesses within decision-making pipelines. Furthermore,  human studies indicate that stage-specific evaluation using \ours correlates more strongly with improvements in user satisfaction than with the final task completion metric alone. Lastly, analyses on combining agents during deployment and the reliability of user simulators highlight promising research directions for future scholars and practitioners.

\section{Background: Bespoke Evaluation of Planning Agents}
 
Recent benchmarks for evaluating agentic tasks adopt rigid, domain-specific metrics to reflect the unique demands of each task setting, as summarized in \autoref{sec:metrics}. This specialization is often necessary: interactive planning tasks vary widely in structure, goals, and interaction modalities, requiring tailored criteria to capture meaningful performance signals.

For instance, AgentBench~\citep{liu2024agentbench} incorporates domain-specific metrics such as Success Rate, Win Rate, F1 Score, and Exact Match to assess agents comprehensively. 
FlowBench~\citep{xiao-etal-2024-flowbench} further enhances the evaluation landscape by adding metrics focused explicitly on tool usage: Tool Invocation, measured by precision, recall, and F1 score for identifying the correct tool configurations; Success Rate, denoting the proportion of entirely successful sessions; and Task Progress, capturing the percentage of goals completed within a session. 
Meanwhile, \taubench~\citep{yao2025taubench} utilizes more direct measures of success with Pass@k, a boolean indicator reflecting success across k attempts, and Pass\textasciicircum{}k, which requires successful outcomes for all k trials.

These metrics are often rigid by design, optimized for narrowly defined outcomes within their respective domains. However, this tight coupling limits their generalizability: it is difficult to apply these metrics across tasks or to capture broader notions of agent capability, especially in open-ended or compositional settings. As noted in prior work~\citep{ribeiro-etal-2020-beyond}, overemphasis on final task success obscures nuanced failures or partial progress, particularly problematic in subjective or exploratory tasks like information aggregation~\citep{gangi-reddy-etal-2025-infogent, yoran-etal-2024-assistantbench}. \textbf{In contrast, \ours seeks to define a more general and domain-agnostic evaluation paradigm that is not tied to narrow success criteria.}

\section{Method}
\label{sec:method}

As a prerequisite step, we set minimum requirements for task planning benchmarks to effectively show realistic scenarios. Specifically, task planning benchmarks should define specific \textbf{tools} (\eg, APIs) and construct domain-specific \textbf{databases}. These assumptions are adopted because their dedicated environmental constraints present challenges in achieving a unified evaluation. 


\begin{wrapfigure}[19]{r}{0.38\textwidth}
    \centering
    \includegraphics[width=0.38\textwidth]{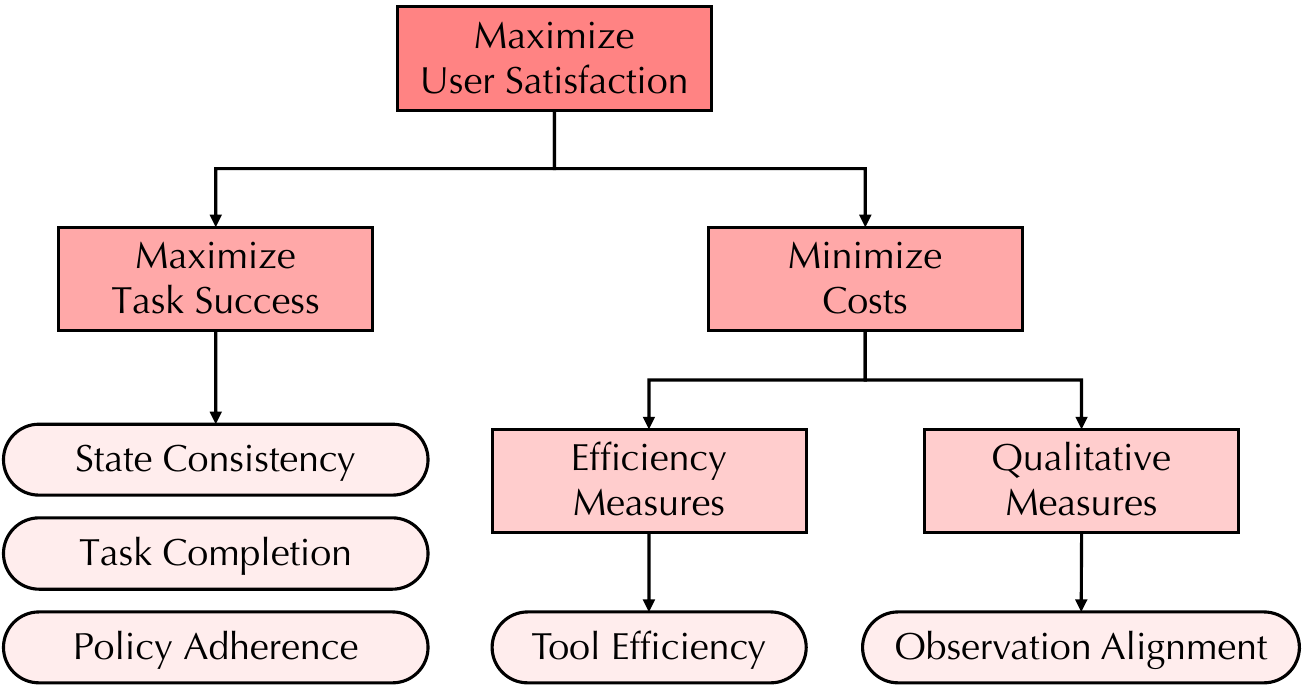}
    \caption{Decision theory-based taxonomy of evaluation metrics in \ours. As discussed in \cite{walker-etal-1997-paradise}, it should be cautious to generalize the original taxonomy to different agents and tasks. Following this, we provide five distinct interpretations of each element, which will be described in \autoref{sec:protocoltask}.}
    \label{fig:decision}
\end{wrapfigure}


\subsection{Metric Design Criteria}

Interactive planning tasks are generally modeled using the POMDP paradigm, where agents devise plans based on partial observational data~\citep{10610981, 10.5555/3692070.3692783, pmlr-v235-xie24j, yao2025taubench}. Building on the framework of \cite{WILLIAMS2007393}, we present an abstract and discrete pipeline consisting of $\mathcal{S}$ (a set of agent states), $\mathcal{A}$ (a set of actions), $\mathcal{O}$ (a set of observations) and $\mathcal{R}$ (a set of  rewards), along with an our additional policy definition $\mathcal{P}$ (a set of global policies).  While task completion is a common performance goal, we hypothesize that it is insufficient: user satisfaction depends on \textit{how} an agent arrives at outcomes~\citep{ASHFAQ2020101473, 10.1145/3477495.3531798}.

To that end, \ours decomposes evaluation into atomic, phase-specific criteria reflecting the multi-stage behavior of LLM agents. These include state consistency, tool efficiency, observation alignment, policy adherence, and task completion, each tied to a specific step in the decision pipeline. Drawing from prior work on decomposed evaluation~\citep{liu-etal-2023-revisiting, saadfalcon2024lmunitfinegrainedevaluationnatural, lee2024checkevalrobustevaluationframework}, we design intuitive LLM-based evaluations for each metric for more consistent and precise diagnostics.

Furthermore, as illustrated in \autoref{fig:decision}, we align \ours with the decision theoretic principles~\citep{walker-etal-1997-paradise}, where disparate metrics collectively estimate user satisfaction. POMDPs provide a natural foundation here, as agents select actions under uncertainty to maximize expected utility, capturing both goal achievement and procedural quality. This integration enables \ours to measure user satisfaction holistically while remaining grounded in theoretical rigor.

Throughout this work, we formulate the basic components of multi-turn, multi-step agents as follows: $T$ denotes the total number of turns (\ie, alternating utterances between the user and the agent) in an interactive session, $M_t$ represents the number of steps in the agent’s internal reasoning process within a single utterance at turn $t$, and $\mathcal{C}_t = \{u_\tau, a_\tau\}_{\tau=1}^t$ signifies the context of user inputs and agent responses observed up to turn $t$.

\subsection{Components of \ours}
\label{sec:protocoltask}

\subsubsection{State Consistency (\texorpdfstring{$\mathcal{S}$}{S})}
\label{sec:stateconsistency}

The intermediate decision-making step of the agent is pivotal in summarizing user requests and determining optimal subsequent actions~\citep{shin-etal-2022-dialogue, wei2022chain}. Given that both pipelined and end-to-end multi-turn interactions are inherently susceptible to error accumulation~\citep{10.1145/3240508.3240605, liao-etal-2021-dialogue}, it is crucial to validate whether intermediate outcomes consistently mediate between user inputs and agent outputs. In this context, the \textit{state consistency} metric measures whether \textbf{an agent correctly aligns user requests with its $k$-th intermediate steps ($z_t^k$).} Notably, these intermediate steps can be represented in either a ``structured format'' (\eg, dialogue states) or ``natural language'' (\eg, Chain-of-Thought). By accommodating both representations, the state consistency can be formulated as follows:

{
\small
\begin{align*}
\text{\ours}_\mathcal{S} = 
 \frac{\displaystyle\sum_{t=1}^{T} \sum_{k=1}^{M_{t}} \text{IsConsistent}\bigl(z_t^k,\;\mathcal{C}_t\backslash \{a_t\}\bigr)}{\displaystyle\sum_{t=1}^{T} M_{t}}
\end{align*}
}

Here, the \texttt{IsConsistent} function serves as a boolean indicator evaluated by LLMs that compares user requests (mostly natural language) with internal states (structured or natural language)\footnote{Practically, as observed by \cite{lee2024checkevalrobustevaluationframework}, evaluating a group of targeted samples collectively, using simple and atomic criteria such as boolean assessments, does not adversely affect performance compared to stepwise evaluation. Therefore, we group $z_t^{1:k}$ in the implementation.}.

\subsubsection{Tool Efficiency (\texorpdfstring{$\mathcal{A}$}{A})}

The management of external tools or functions, typically in the form of APIs, incurs operational costs and affects the task performance, underscoring the importance of evaluating their effective utilization. In particular, the occurrence of failed API calls prior to collecting complete information from users can result in unnecessary expenditures of resources (\ie, time and money) and negative user experience. To address this, the \textit{tool efficiency} metric first considers \textbf{the total number of tool calls ($N_T$)}. For a given task, agent scenarios that require fewer API calls are considered indicative of more efficient API management.

Additionally, due to the stochastic nature of natural language prompting, generating well-structured API calls poses a significant challenge for certain agents. Even when an agent invokes an API at an appropriate time, iterative attempts due to incomplete generation will eventually lead to additional, avoidable costs. Consequently, this metric also calculates \textbf{the number of failed tool generation attempts ($N_F$)}. Formally, we propose the following tool efficiency measure:

{
\small
\begin{align*}
\text{\ours}_\mathcal{A} = \frac{N_T - N_F}{N_T + N_F}
\end{align*}
}

where the numerator rewards successful calls and the denominator penalizes excessive or failed calls. 
A higher value of $\text{\ours}_\mathcal{A}$ indicates more efficient tool usage: maximizing successful calls while minimizing total and failed calls. 

\subsubsection{Observation Alignment (\texorpdfstring{$\mathcal{O}$}{O})}
\label{sec:obsalignment}

The \textit{observation alignment} metric evaluates \textbf{whether observations appearing within the context align with what the user requires.} Specifically, it is calculated with a boolean criteria for each observation (\ie, retrieved database entity), preceded by capturing the number of observations ($|O|$) explicitly appear within the agent responses. Considering the fact that both user utterance and agent response are represented in natural languages, it can be measured via a set of atomic LLM evaluations:

{
\small
\begin{align*}
\text{\ours}_\mathcal{O} =
\frac{1}{|T_{\text{obs}}|} \sum_{t \in T_{\text{obs}}} \left(\frac{1}{|O^{(t)}|} \sum_{o \in O^{(t)}} \text{IsAligned}(o, \mathcal{C}_t\backslash \{a_t\})\right) \,, \nonumber  \text{where}\; T_{obs} = \{t \in T : |O^{(t)}| > 0\}
\end{align*}
}

Intuitively, $T_{obs}$ denotes a set of turns where observations are present within agent responses ($|O|>0$), and observations of a specific turn $t\in T_{obs}$ are extracted from LLM (denoted $O^{(t)}$). In addition, \texttt{IsAligned} serves as a boolean indicator assessing whether each observation aligns with conversational context. By quantifying and improving observation alignment, we not only promote the clarity of agent outputs but also minimize the system's overall costs by decreasing the possibility of repetitively calling tools.

\subsubsection{Policy Alignment (\texorpdfstring{$\mathcal{P}$}{P})}

Interactive agent benchmarks assume certain policies (\ie, a behavioral rule appearing in agent responses, not only emerging in their internal states) that are globally reflected across the interactive sessions~\citep{xiao-etal-2024-flowbench}, mostly in a form of system prompt. 

The \textit{Policy alignment} is a session-level metric measuring whether a predefined set of policies ($\mathcal{P}$) are consistently followed throughout interactive sessions.

{
\small
\begin{align*}
\text{\ours}_\mathcal{P} = \frac{1}{|\mathcal{P}|} \sum_{p \in \mathcal{P}}{\text{IsAdherent}(p; \mathcal{C}_T)} 
\end{align*}
}

Similar to the state consistency~(\autoref{sec:stateconsistency}) and observation alignment~(\autoref{sec:obsalignment}) metrics, \texttt{IsAdherent} serves as an LLM-evaluated boolean indicator deciding consistency between each policy and interaction context.

\begin{table*}[t]
\centering \small
\renewcommand{\arraystretch}{0.4}
\setlength{\tabcolsep}{11pt} 
\caption{Benchmark statistics. We use a validation set for \benchone that demonstrates a similar performance pattern with the test set.}
\begin{tabular*}{\textwidth}{cccl} 
\toprule
 & \textbf{\# of Scenarios} & \textbf{\# of Tools} & \textbf{\# of Database} \\ \midrule[1pt]
\benchone~\cite{pmlr-v235-xie24j}  & 180  & 7 & 3,865,195 total, 3,827,361 max for a tool \\ \midrule
\benchtwo~\cite{yao2025taubench}   & 50  & 13 & 500 users, 300 flights, 2,000 reservations \\ \midrule
\benchthree~\cite{yao2025taubench} & 115 & 15 & 500 users, 50 products, 1,000 orders \\
\bottomrule[1pt]
\end{tabular*}
\label{tab:statsbench}
\end{table*}

\subsubsection{Task Completion (\texorpdfstring{$\mathcal{R}$}{R})}

As in many agent studies, the primary objective of a task agent is to effectively accomplish goal-oriented tasks. The \textit{task completion} metric adheres to the task-specific performance criteria established by each benchmark, since benchmark scenarios typically define their own evaluation frameworks to determine what constitutes a ``completed task.'' Specifically, most metrics in diverse benchmarks demonstrated in \autoref{tab:metrics} can be regarded as task completion metrics since they mainly focus on whether agents successfully accomplish the given task. Again, it is important to note that \textbf{\ours incorporates existing task completion metrics} to complement the holistic evaluation, rather than excluding them.

\subsubsection{Agent Interaction Pattern}
\label{sec:interactionpattern}

While prior works have typically reported on dataset-level statistics (\eg, basic information in \autoref{tab:statsbench}), it has rarely addressed how agents actually interact within their environments. We additionally report the following information, providing an overview of the interactive tendency of employed agents in their specific scenarios.

\paragraph{The Number of Turns:} Different from offline dialogue datasets where a deterministic number of turns is provided, recently employed online evaluation scenarios typically omit details about the agent interactivity. However, the number of interactive turns serves as a critical indicator of the conversational or interactive dynamics. In general, planning scenarios with more turns tend to require further management of user constraints, suggesting the need for sustained engagement and state tracking. 

\paragraph{The Number of Steps:} Steps refer to internal actions or reasoning hops, such as goal decomposition or tool use. This feature reflects the agent's ability to handle complex tasks that require breaking down overarching goals into smaller, manageable actions. Fewer steps may suggest shallow or generic reasoning, while more steps indicate finer control. Correlating the count of steps with performance and the length of interaction can reveal whether agents are making meaningful progress or getting stuck in unproductive loops.


\section{Experiments}

\subsection{Benchmarks}

To verify our hypothesis on the evaluation following behavioral stages, we employ challenging task planning benchmarks and compare agent performance based solely on task completion with that achieved using \ours. Specifically, we utilize three task domains of two benchmarks: TravelPlanner~\citep{pmlr-v235-xie24j}, a single-turn dataset in the itinerary domain, and \taubench~\citep{yao2025taubench}, a multi-turn dataset encompassing airline and retail domains. \autoref{tab:statsbench} presents the statistics of these benchmarks. These benchmarks are selected due to their rich environmental constraints that conform to the prerequisite assumptions described in \autoref{sec:method} (\ie, predefined tools and external database). For the evaluation metric for task completion ($\mathcal{R}$), pass rate is used for \benchone, and pass{\char94}\textit{k} is used in \taubench. Refer to \autoref{tab:metrics} for the description of each metric.

\subsection{LLMs for Agent, User Simulator, and Task Evaluator}
\begin{table*}[t]
\centering \small
\renewcommand{\arraystretch}{0.4}
\caption
{
\ours evaluation result. The relative performance ranking among agents for each metric is differentiated with colors (\ie, darker color indicates more competitive performance; different colors are applied for proprietary and open-source models). Best average performance is indicated in \textbf{bold}, and second-best is \underline{underlined}. 
}
\setlength{\tabcolsep}{2.8pt} 
\begin{tabular*}{\textwidth}{c!{\vrule}cccccc!{\vrule}cccccc!{\vrule}cccccc} 
\toprule
 & \multicolumn{6}{c}{\benchone}  & \multicolumn{6}{!{\vrule}c!{\vrule}}{\benchtwo} & \multicolumn{6}{c}{\benchthree} \\ \midrule
Agent & $\mathcal{S}$ & $\mathcal{A}$ & $\mathcal{O}$ & $\mathcal{P}$ & $\mathcal{R}$ & {\scriptsize AVG.} &
$\mathcal{S}$ & $\mathcal{A}$ & $\mathcal{O}$ & $\mathcal{P}$ & $\mathcal{R}$ & {\scriptsize AVG.} &
$\mathcal{S}$ & $\mathcal{A}$ & $\mathcal{O}$ & $\mathcal{P}$ & $\mathcal{R}$ & {\scriptsize AVG.} \\ \midrule[1pt]


\multicolumn{19}{c}{\textbf{\textit{Proprietary Models}}} \\ \midrule[1pt]

\fouro & 
\cellcolor{\colorprop!\ranktwo}.98 & 
\cellcolor{\colorprop!\rankfour}.79 & 
\cellcolor{\colorprop!\rankfour}.01 & 
\cellcolor{\colorprop!\rankthree}.48 & 
\cellcolor{\colorprop!\rankfour}.01 & 
.44 &

\cellcolor{\colorprop!\ranktwo}.58 & 
\cellcolor{\colorprop!\ranktwo}.96 & 
\cellcolor{\colorprop!\ranktwo}.73 & 
\cellcolor{\colorprop!\rankfour}.83 &
\cellcolor{\colorprop!\ranktwo}.42 & 
\underline{.70} &

\cellcolor{\colorprop!\rankthree} .53 & 
\cellcolor{\colorprop!\ranktwo}.93 & 
\cellcolor{\colorprop!\ranktwo}.79 & 
\cellcolor{\colorprop!\ranktwo}.89 & 
\cellcolor{\colorprop!\ranktwo}.54 & 
\underline{.74} \\ \midrule

\fouromini & 
\cellcolor{\colorprop!\rankfour}.94 & 
\cellcolor{\colorprop!\ranktwo}.95 & 
\cellcolor{\colorprop!\rankfive}.00 & 
\cellcolor{\colorprop!\ranktwo}.50 & 
\cellcolor{\colorprop!\rankfive}.00 & 
\underline{.48} &

\cellcolor{\colorprop!\rankfive}.49 & 
\cellcolor{\colorprop!\rankthree}.92 & 
\cellcolor{\colorprop!\rankthree}.71 & 
\cellcolor{\colorprop!\ranktwo}.84 & 
\cellcolor{\colorprop!\rankfour}.26 & 
.64 &

\cellcolor{\colorprop!\rankfour}.48 & 
\cellcolor{\colorprop!\rankthree}.89 & 
\cellcolor{\colorprop!\rankthree}.72 & 
\cellcolor{\colorprop!\ranktwo}.89 &
\cellcolor{\colorprop!\rankthree}.46 & 
.69 \\ \midrule

\threeturbo & 
\cellcolor{\colorprop!\rankthree}.95 & 
\cellcolor{\colorprop!\rankthree}.93 & 
\cellcolor{\colorprop!\rankfive}.00 & 
\cellcolor{\colorprop!\ranktwo}.50 & 
\cellcolor{\colorprop!\rankfive}.00 & 
\underline{.48} &

\cellcolor{\colorprop!\rankthree}.56 & 
\cellcolor{\colorprop!\rankfive}.79 & 
\cellcolor{\colorprop!\rankfive}.65 & 
\cellcolor{\colorprop!\rankfive}.73 & 
\cellcolor{\colorprop!\rankfive}.08 & 
.56 &

\cellcolor{\colorprop!\rankfive}.44 & 
\cellcolor{\colorprop!\rankfive}.69 & 
\cellcolor{\colorprop!\rankfive}.57 & 
\cellcolor{\colorprop!\rankfive}.78 &
\cellcolor{\colorprop!\rankfour}.21 & 
.54 \\ \midrule

\gemini & 
\cellcolor{\colorprop!\rankone}.99 & 
\cellcolor{\colorprop!\rankfive}.71 & 
\cellcolor{\colorprop!\rankfive}.00 & 
\cellcolor{\colorprop!\rankfour}.31 & 
\cellcolor{\colorprop!\rankfive}.00 & 
.40 &

\cellcolor{\colorprop!\rankfour}.55 & 
\cellcolor{\colorprop!\rankfour}.84 & 
\cellcolor{\colorprop!\rankfour}.66 & 
\cellcolor{\colorprop!\rankone}{.87} & 
\cellcolor{\colorprop!\rankthree}.32 & 
.65 &

\cellcolor{\colorprop!\ranktwo}.60 & 
\cellcolor{\colorprop!\rankfour}.85 & 
\cellcolor{\colorprop!\rankthree}.72 & 
\cellcolor{\colorprop!\rankthree}.86 &
\cellcolor{\colorprop!\rankfive}.15 & 
.64 \\ \midrule

\sonnet & 
\cellcolor{\colorprop!\rankone}.99 & 
\cellcolor{\colorprop!\rankone}.99 & 
\cellcolor{\colorprop!\rankfour}.01 & 
\cellcolor{\colorprop!\rankone}.56 & 
\cellcolor{\colorprop!\rankfour}.01 & 
\textbf{.51} &

\cellcolor{\colorprop!\rankone}.80 & 
\cellcolor{\colorprop!\rankone}.97 & 
\cellcolor{\colorprop!\rankone}.84 & 
\cellcolor{\colorprop!\rankone}.87 & 
\cellcolor{\colorprop!\rankone}.46 & 
\textbf{.79} &

\cellcolor{\colorprop!\rankone}.73 & 
\cellcolor{\colorprop!\rankone}.94 & 
\cellcolor{\colorprop!\rankone}.85 & 
\cellcolor{\colorprop!\rankone}.91 &
\cellcolor{\colorprop!\rankone}.62 & 
\textbf{.81} \\ \midrule[1pt]


\multicolumn{19}{c}{\textbf{\textit{Open-Source Models}}} \\ \midrule[1pt]

\mlarge & 
\cellcolor{\coloros!\rankfour}.87 & 
\cellcolor{\coloros!\rankthree}.77 & 
\cellcolor{\coloros!\rankfive}.00 & 
\cellcolor{\coloros!\ranktwo}.36 & 
\cellcolor{\coloros!\rankfive}.00 & 
.40 &

\cellcolor{\coloros!\rankthree}.31 & 
\cellcolor{\coloros!\rankone}.96 & 
\cellcolor{\coloros!\rankthree}.50 & 
\cellcolor{\coloros!\rankthree}.76 & 
\cellcolor{\coloros!\rankthree}.26 & 
\underline{.56} &

\cellcolor{\coloros!\rankthree}.23 & 
\cellcolor{\coloros!\rankone}.91 & 
\cellcolor{\coloros!\ranktwo}.44 & 
\cellcolor{\coloros!\ranktwo}.80 &
\cellcolor{\coloros!\rankthree}.34 & 
.54 \\ \midrule

\mixtral & 
\cellcolor{\coloros!\rankthree}.94 & 
\cellcolor{\coloros!\rankthree}.77 & 
\cellcolor{\coloros!\rankfive}.00 & 
\cellcolor{\coloros!\rankthree}.27 &  
\cellcolor{\coloros!\rankfive}.00 & 
.40 &

\cellcolor{\coloros!\rankfour}.25 & 
\cellcolor{\coloros!\rankthree}.85 & 
\cellcolor{\coloros!\rankfour}.17 & 
\cellcolor{\coloros!\rankfour}.64 & 
\cellcolor{\coloros!\ranktwo}.28 & 
.44 &

\cellcolor{\coloros!\ranktwo}.27 & 
\cellcolor{\coloros!\rankthree}.50 & 
\cellcolor{\coloros!\rankthree}.27 & 
\cellcolor{\coloros!\rankfour}.65 & 
\cellcolor{\coloros!\rankthree}.05 & 
.35 \\ \midrule

\llama & 
\cellcolor{\coloros!\ranktwo}.95 & 
\cellcolor{\coloros!\ranktwo}.96 & 
\cellcolor{\coloros!\rankfive}.00 & 
\cellcolor{\coloros!\rankone}.39 & 
\cellcolor{\coloros!\rankfive}.00 & 
\underline{.46} &

\cellcolor{\coloros!\rankone}.38 & 
\cellcolor{\coloros!\ranktwo}.90 & 
\cellcolor{\coloros!\rankone}.70 & 
\cellcolor{\coloros!\rankone}.81 &
\cellcolor{\coloros!\rankone}.30 & 
\textbf{.62} &

\cellcolor{\coloros!\rankone}.32 & 
\cellcolor{\coloros!\ranktwo}.88 & 
\cellcolor{\coloros!\rankone}.79 & 
\cellcolor{\coloros!\rankone}.87 &
\cellcolor{\coloros!\ranktwo}.36 & 
\textbf{.64} \\ \midrule

\qwen & 
\cellcolor{\coloros!\rankone}.98 & 
\cellcolor{\coloros!\rankone}.97 & 
\cellcolor{\coloros!\rankfive}.00 & 
\cellcolor{\coloros!\rankone}.39 & 
\cellcolor{\coloros!\rankfive}.00 & 
\textbf{.47} &

\cellcolor{\coloros!\ranktwo}.35 & 
\cellcolor{\coloros!\ranktwo}.90 & 
\cellcolor{\coloros!\ranktwo}.58 & 
\cellcolor{\coloros!\ranktwo}.79 & 
\cellcolor{\coloros!\rankfour}.20 & 
\underline{.56} &

\cellcolor{\coloros!\rankfour}.27 & 
\cellcolor{\coloros!\rankfour}.75 & 
\cellcolor{\coloros!\ranktwo}.53 & 
\cellcolor{\coloros!\rankthree}.84 &
\cellcolor{\coloros!\rankone}.38 & 
\underline{.55} \\

\bottomrule[1pt]
\end{tabular*}
\label{tab:task}

\end{table*}

We test diverse task planning agents consisting of five proprietary models (\fouro, \fouromini, \threeturbo, \geminifull, \sonnet) and four open-weight models (\mlarge\footnote{123B-sized model: \url{https://huggingface.co/mistralai/Mistral-Large-Instruct-2411}}, \mixtral, \llama, \qwen), with a temperature of 0.0, respectively. Since \taubench requires multi-turn interactions between a user and agent, we employ \fouro as a user simulator with the same instruction as the original work~\citep{yao2025taubench}. For the LLM evaluator leveraged in measuring \ours metrics, we employ \llama~\citep{grattafiori2024llama3herdmodels}\footnote{We confirm \llama shows better performance compared to \fouromini through our preliminary investigation.}. All prompts used in our experiments are listed in \autoref{sec:prompts}.

\begin{table*}[t]
\centering \small
\caption{
The average number of turns and steps within a single interactive session (the agent interaction pattern defined in \autoref{sec:interactionpattern}). Note that \texttt{Avg. Steps} denotes the number of internal agentic processes in each turn not visible in superficial interactions.
}
\renewcommand{\arraystretch}{0.4}
\setlength{\tabcolsep}{6pt} 
\begin{tabular*}{\textwidth}{c!{\vrule}cc!{\vrule}cc!{\vrule}cc} 
\toprule
 & \multicolumn{2}{c}{\benchone}  & \multicolumn{2}{!{\vrule}c!{\vrule}}{\benchtwo} & \multicolumn{2}{c}{\benchthree} \\ \midrule
 Agent & Avg. Turns & Avg. Steps & Avg. Turns & Avg. Steps & Avg. Turns & Avg. Steps \\ \midrule[1pt]

 \multicolumn{7}{c}{\textbf{\textit{Proprietary Models}}} \\ \midrule[1pt]

\fouro & 1.00$_{\pm0.00}$ & 18.71$_{\pm6.91}$ & 7.70$_{\pm 3.08}$ & 0.93$_{\pm1.71}$ & 8.26$_{\pm2.54}$ & 0.98$_{\pm1.23}$  \\ \midrule

\fouromini & 1.00$_{\pm0.00}$ & 15.16$_{\pm6.67}$ & 8.72$_{\pm3.77}$ & 0.81$_{\pm1.79}$ & 8.52$_{\pm3.09}$ & 1.07$_{\pm1.50}$  \\ \midrule

\threeturbo  & 1.00$_{\pm0.00}$ & 12.87$_{\pm6.44}$ & 7.66$_{\pm3.70}$ & 0.94$_{\pm1.12}$ & 7.97$_{\pm2.50}$ & 1.17$_{\pm1.19}$  \\ \midrule

\gemini & 1.00$_{\pm0.00}$ & 16.57$_{\pm6.97}$ & 3.04$_{\pm2.14}$ & 0.47$_{\pm0.70}$ & 6.74$_{\pm4.77}$ & 0.44$_{\pm0.68}$  \\ \midrule

\sonnet  & 1.00$_{\pm0.00}$ & 18.36$_{\pm6.84}$ & 6.20$_{\pm2.10}$ & 1.03$_{\pm1.58}$ & 7.71$_{\pm2.06}$ & 1.15$_{\pm1.49}$  \\ \midrule[1pt]

\multicolumn{7}{c}{\textbf{\textit{Open-Weight Models}}} \\ \midrule[1pt]

\small \mlarge  & 1.00$_{\pm0.00}$ & 14.16$_{\pm7.48}$ & 8.46$_{\pm6.33}$ & 0.62$_{\pm1.74}$ & 12.51$_{\pm6.44}$ & 0.68$_{\pm0.96}$  \\ \midrule

\small \mixtral & 1.00$_{\pm0.00}$ & 18.44$_{\pm7.84}$ & 18.14$_{\pm9.99}$ & 0.01$_{\pm0.12}$ & 16.11$_{\pm9.69}$ & 0.01$_{\pm0.09}$  \\ \midrule

\small \llama & 1.00$_{\pm0.00}$ & 18.52$_{\pm6.54}$ & 8.02$_{\pm4.33}$ & 1.22$_{\pm1.63}$ & 6.44$_{\pm4.03}$ & 1.31$_{\pm1.31}$  \\ \midrule

\small \qwen & 1.00$_{\pm0.00}$ & 18.20$_{\pm7.36}$ & 18.74$_{\pm9.21}$ & 0.04$_{\pm0.86}$ & 17.10$_{\pm9.73}$ & 0.01$_{\pm0.09}$  \\

\bottomrule[1pt]
\end{tabular*}
\label{tab:turnstep}
\end{table*}


\subsection{Results}

\paragraph{Behavioral Stage Diagnosis:} \autoref{tab:task} presents the evaluation results following \ours (see \autoref{sec:failure} for agents' erroneous behaviors at each stage). We differentiate the color ranking of proprietary and open-weight models to visually observe patterns. Along with individual metrics in \ours, we report the average score to see the overall behavioral performance.

A notable observation is that \textbf{the conventional task completion metric ($\mathcal{R}$) does not necessarily reflect the performance of intermediate phases.} For instance, while \qwen achieves the highest performance on task completion (.38) among open-weight models in \benchthree, another \ours results lag behind other models (\eg, \llama). Moreover, although most agents in \benchone show the same zero task completion performance\footnote{This low task completion result is consistent with the original paper's results~\citep{pmlr-v235-xie24j}.}, each model has different intermediate capability patterns according to \ours metrics. These observed patterns imply a potential discrepancy between task completion performance values and qualitative user preferences, as user satisfaction is influenced not only by task completion but also by the quality of interactions across all phases. We introduce a human study for verifying this discrepancy in \autoref{sec:relationship}.  

\paragraph{Interaction Pattern Diagnosis:}

We summarize each agent's interaction pattern in \autoref{tab:turnstep}. While \benchone is characterized by single-turn interactions with numerous intermediate steps prior to producing a final response, \taubench exhibits fewer intermediate steps per turn but involves multiple turns. These patterns reveal certain behavioral tendencies among the agents; for instance, \textbf{competitive agents (\eg, \sonnet, \llama) appear to engage in more extensive internal reasoning within individual turns, yet participate in fewer turns overall in multi-turn benchmarks.} This observation provides insightful suggestions on scenario diversification. For example, when adapting \benchone for multi-turn scenarios, careful consideration must be given to balancing the number of turns and steps within each turn. Given that service providers often target different application scenarios, analyzing interaction patterns across benchmarks can inform the strategic selection of those most aligned with specific deployment contexts.

\section{Analyses and Discussions}
\label{sec:analysis}

\subsection{Relationship Between Task Completion, \ours, and User Satisfaction}
\label{sec:relationship}

Our hypothesis regarding the results presented in \autoref{tab:task} is that factors beyond task completion may have significantly influenced overall user satisfaction. To investigate this through a human study, we recruit 16 graduate-level participants, a sample size congruent to prior human-computer interaction research~\cite{10.1145/1735223.1735255, 10.1145/2858036.2858498}. This study is conducted using two controlled scenarios, described as follows:

\paragraph{(1) Same \texorpdfstring{$\mathcal{R}$}{R} \& Different {\small AVG.}}

We select \fouromini and \mlarge in \benchtwo, as these two models exhibit identical task completion performance (.26), yet differ in their \ours average scores (.64 and .56, respectively).

\paragraph{(2) Same {\small AVG.} \& Different \texorpdfstring{$\mathcal{R}$}{R}}

We select \gemini and \llama in \benchthree, as these two models exhibit identical \ours average score (.64), yet differ in their task completion performance (.15 and .36, respectively).

\begin{wrapfigure}[25]{r}{0.4\textwidth}
    \centering
    \vspace{8mm}
    \includegraphics[width=0.4\textwidth]{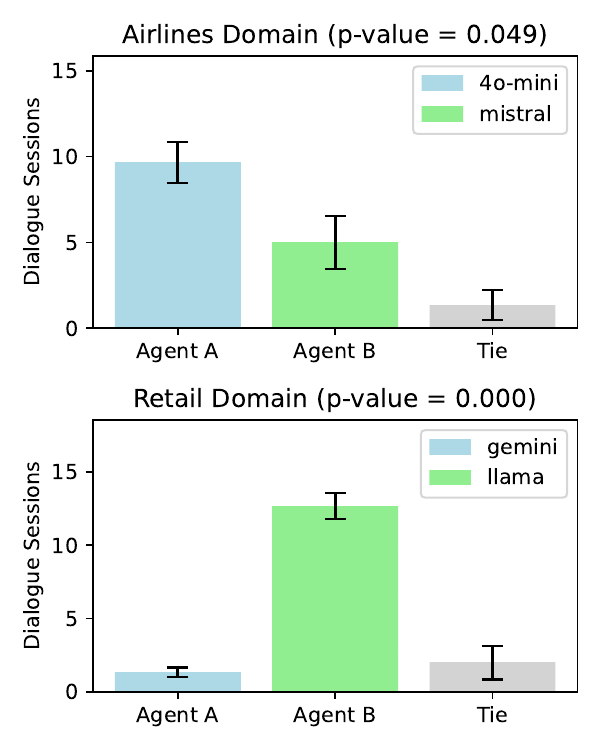}
    \caption{Human study results examining preferred interactions.}
    \label{fig:human}
\end{wrapfigure}


Participants are presented with two multi-turn conversations generated by different agents interacting with a \texttt{gpt-4o} user simulator. They are asked to indicate which conversation they find more satisfactory. If both conversations are perceived as equally (un)satisfactory, participants are given the option to select ``no preference.'' Each participant evaluates a total of six conversation pairs (randomly sampled three pairs of conversations, respectively). Detailed instructions are demonstrated in \autoref{sec:guideline}, and we provide qualitative analysis for participants' decisions in \autoref{sec:justification}.
\begin{table*}[t]
\centering \small
\caption
{
Selected results for mixing agent experiments in \benchtwo. Agents indicated with ``\textbf{interm.}'' are utilized in an intermediate understanding component. Full results and additional discussions are presented in \autoref{tab:mixfull} and \autoref{sec:mixing}.
}
\renewcommand{\arraystretch}{1.0}
\begin{tabular}{l!{\vrule}cccc>{\columncolor{lightgray!30}}c>{\columncolor{lightgray!30}}c} 
\toprule
& $\mathcal{S}$ & $\mathcal{A}$ & $\mathcal{O}$ & $\mathcal{P}$ & $\mathcal{R}$ & {\scriptsize AVG.}\\ \midrule[1pt]

\mlarge & 
.383$_{\pm0.038}$  & 
.924$_{\pm0.046}$ & 
.581$_{\pm0.022}$ & 
.777$_{\pm0.005}$ & 
\textbf{.358}$_{\pm0.029}$ & 
.605
\\

\quad + \qwen \textbf{interm.} & 
.417$_{\pm0.031}$ & 
.939$_{\pm0.017}$ & 
.545$_{\pm0.029}$ & 
.809$_{\pm0.008}$ & 
.336$_{\pm0.050}$ & 
\textbf{.609}
\\

\midrule

\qwen & 
.403$_{\pm0.021}$  & 
.951$_{\pm0.015}$ & 
.637$_{\pm0.031}$ & 
.821$_{\pm0.008}$ & 
{.329}$_{\pm0.021}$ & 
{.628}
\\

\quad + \llama \textbf{interm.} & 
.447$_{\pm0.020}$ & 
.958$_{\pm0.014}$ & 
.663$_{\pm0.022}$ & 
.831$_{\pm0.008}$ & 
.272$_{\pm0.051}$ & 
\textbf{.634}
\\

\quad + \mlarge \textbf{interm.} & 
.423$_{\pm0.039}$ & 
.933$_{\pm0.032}$ & 
.578$_{\pm0.050}$ & 
.776$_{\pm0.009}$ & 
\textbf{.367}$_{\pm0.013}$ & 
.615
\\

\midrule

\llama & 
.463$_{\pm0.055}$  & 
.952$_{\pm0.013}$ & 
.643$_{\pm0.035}$ & 
.818$_{\pm0.015}$ & 
.263$_{\pm0.056}$ & 
\textbf{.628}
\\

\quad + \qwen \textbf{interm.} & 
.404$_{\pm0.033}$ & 
.953$_{\pm0.007}$ & 
.599$_{\pm0.016}$ & 
.803$_{\pm0.014}$ & 
\textbf{.361}$_{\pm0.019}$ & 
.624
\\

\bottomrule[1pt]
\end{tabular}
\label{tab:mix}
\end{table*}

The results of the human evaluation are presented in \autoref{fig:human}. For each scenario, we conduct binomial tests by excluding ``no preference'' responses, yielding statistically significant results (p-value $<$ 0.05 in both cases). In study (1), we observe that participants tend to \textbf{favor conversations with higher average score when task completion is held constant}. Conversely, in study (2), \textbf{task completion still remains a strong predictor of user satisfaction when the average score is identical}. These findings highlight the complementary roles of general task completion and \ours metrics in capturing different facets of user satisfaction, as demonstrated in \autoref{tab:task}. Our study thus supports the use of \ours as a meaningful proxy for user satisfaction and motivates future research into evaluation frameworks that go beyond simple task outcomes.

\subsection{Can Mixing Agents Lead to Better Performance?}
\label{sec:mixing}

Building upon the insights from previous experiments, we are further motivated to explore the potential for enhancing performance and user satisfaction by leveraging diverse agents strengths within the agentic pipeline. To facilitate this investigation, we separate the agent's interactive process into two distinct components: (1) intermediate understanding and (2) response generation. This separation enables targeted improvements, wherein state consistency and tool efficiency are addressed in the first stage, whereas observation alignment and policy adherence are the focus of the second. Specifically, we employ one agent to understand user requests and call proper tools, then replace it with the other when starting to generate responses.

For faster investigation of this experimental study, we run FP8-quantized versions of three open-weight models (\llama, \qwen, \mlarge), alongside \llama as the user simulator, and evaluate on \benchtwo with five repeated runs per configuration. Selected results are presented in \autoref{tab:mix}. \textbf{While acknowledging that arbitrary combinations of agents do not consistently yield improved performance, even leading to drop in performance, certain pairings demonstrate notable enhancements in either task completion or the average \ours score}. Notably, combining \qwen (for intermediate understanding) with \llama (for response generation) leads to a substantial improvement in task completion (.263 $\rightarrow$ .361) without compromising the average \ours score too much (.628 $\rightarrow$ .624). These findings suggest a promising direction for future research to strategically arrange agents across different stages of the task planning pipeline. For additional details and comprehensive experimental results beyond the scope of this discussion, please refer to \autoref{sec:mixingappendix}.

\subsection{Analysis on the Reliability of User Simulator}
\label{sec:usersim}

To highlight potential limitations of multi-turn evaluation tasks, including those in our own study, we conduct a thorough analysis of user-agent conversations from \benchtwo, where \fouro served as the user simulator. We manually examine all user utterances against corresponding user instructions to assess whether the user's behavior aligns with the intended instructions.

Our analysis reveals that in \textbf{11 out of 50 conversations (22\%), the user simulator demonstrates behaviors not in line with its instructions}. While these deviations do not necessarily lead to failures in task completion, they highlight the importance of considering user simulator performance when evaluating an agent. Although addressing this limitation is left for future work, we identify the following erroneous patterns where a user simulator does not adhere to its instructions: (1) proactivity, (2) instruction contradiction, (3) missing details, and (4) misinterpretation. A comprehensive description of these error categories is provided in \autoref{sec:usererror}.

\section{Conclusion}


We propose \ours, a task-agnostic evaluation framework based on a partially observable Markov Decision Process (POMDP), to more accurately capture how user satisfaction emerges from both intermediate behaviors and final outcomes in interactive task planning. Unlike traditional metrics that focus solely on task completion, \ours provides a holistic lens for assessing agent performance by modeling the user's evolving experience throughout the interaction. Our analysis reveals that different models excel at different stages of a task, underscoring the limitations of evaluating performance through a single end-point measure. Human studies further demonstrate that stage-specific behavioral indicators often correlate more strongly with overall user satisfaction than task completion alone. These findings highlight the importance of comprehensive evaluation methods and establish \ours as a principled, diagnostic tool for developing more transparent, robust, and user-aligned planning agents.

\section*{Limitations and Future Directions}

While we introduce \ours as a diagnostic framework for estimating user satisfaction, we acknowledge its limitations that offer avenues for future research. First, \ours relies on LLMs to serve as the evaluator for its atomic criteria. Although the assessment employs a model based on boolean criteria, demonstrated to be more reliable than long-text evaluations, the accuracy and consistency of the results may nonetheless depend on the specific capabilities and potential bias of the selected model. Furthermore, as discussed in \autoref{sec:usersim}, ensuring the reliability and realism of user simulators remains an open challenge, warranting further investigation. As a future direction, we believe that \ours can provide fine-grained rewards to enhance human-LLM collaboration, thereby further improving user satisfaction and demonstrating synergies with collaborative simulation~\cite{wu2025collabllm}.

\bibliography{reference}
\bibliographystyle{plain} 

\clearpage
\appendix
\onecolumn
\renewcommand{\sectionautorefname}{Appendix}

\section{Evaluation Metrics Across Task Planning Benchmarks}
\label{sec:metrics}
\begin{table*}[ht]
\centering
\caption
{
Performance evaluation metrics of service planning benchmarks. Note that metrics sharing the same title (\eg, Success Rate) have distinct definitions across different benchmarks.
}
\renewcommand{\arraystretch}{1.0}
\setlength{\tabcolsep}{8pt} 
\begin{tabular*}{\textwidth}{ccl} 
\toprule
\textbf{Benchmark}  & \textbf{Metric} & \textbf{Description} \\ \midrule[1pt]

    \multicolumn{3}{c}{\textbf{\textit{Multi-Turn Planning Benchmarks}}}\\ \midrule
   
   \multirow{2}{*}{MultiWOZ~\cite{budzianowski-etal-2018-multiwoz}} & Success Rate & The system answered all requested attributes. \\ 
    & Inform Rate & The system has provided an appropriate entity. \\ \midrule

    \multirow{2}{*}{AgentBench~\cite{liu2024agentbench}} & \multirow{2}{*}{Domain-Specific} & Different metrics such as Success Rate, Win Rate, \\ 
    & & F1 Score, Exact Match, etc. are adopted. \\ \midrule

    \multirow{3}{*}{WebLINX~\cite{10.5555/3692070.3693410}} & Intent Match & Boolean indicator for correct intent prediction. \\ 
    & Element Similarity & Correctly predicted function arguments. \\ 
    & Text Similarity & Lexical similarity of arguments across functions. \\ \midrule

    \multirow{3}{*}{FlowBench~\cite{xiao-etal-2024-flowbench}} & Tool Invocation & Correctly identified tool configs (P/R/F1).  \\ 
    & Success Rate & Proportion of completely successful sessions. \\ 
    & Task Progress & Percentage of completed goals within a session. \\ \midrule

    \multirow{2}{*}{$\tau$-Bench~\cite{yao2025taubench}} & Pass@\textit{k} & Boolean success indicator for \textit{k} attempts. \\ 
    & Pass{\char94}\textit{k} & Whether all \textit{k} trials are successful. \\ \midrule
    
    \multicolumn{3}{c}{\textbf{\textit{Single-Turn Planning Benchmarks}}}\\ \midrule

    \multirow{2}{*}{WebShop~\cite{NEURIPS2022_82ad13ec}} & Task Score & The average reward across episodes. \\ 
    & Success Rate & The proportion of fully rewarded instructions. \\ \midrule

    \multirow{2}{*}{TravelPlanner~\cite{pmlr-v235-xie24j}} & Delivery Rate & The agent completed tasks within limited steps. \\ 
    & Pass Rate & The agent satisfied all plans and constraints. \\

\bottomrule[1pt]
\end{tabular*}
\label{tab:metrics}
\end{table*}

\newpage

\section{Prompts}
\label{sec:prompts}

Basically, we keep the same agent prompt as the original paper~\citep{pmlr-v235-xie24j, yao2025taubench}. The following prompts are evaluation prompts adopted in \ours evaluation.

\begin{table}[ht!]
    \centering\small\ttfamily
    \def\arraystretch{1.4}
    \setlength{\tabcolsep}{0.5em}
    \begin{tabularx}{\columnwidth}{|>{\raggedright}X|}
        \hline
        \rowcolor{tableheader}\textbf{\textsf{LLM Evaluation Prompt for State Consistency}} \tabularnewline
        \arrayrulecolor{black}\hline
        \small 
        \textbf{Instruction:} You are tasked with evaluating whether each agent's \textbf{intermediate state} (which can be either a \textbf{thought}-the agent's internal reasoning-or an \textbf{action}-an API call) accurately reflects and mediates between: \\
        1. The user's requests (in the dialogue so far). \\
        2. Any previously established agent states. \\
        
        Your evaluation should focus on whether the agent's intermediate steps exhibit clear, consistent reasoning that aligns the user's inputs with the agent's outputs, without introducing errors or contradicting earlier information. \\

        \textbf{Evaluation Criteria:} \\
        1. \textbf{Consistency with the user request} \\
        \quad - Does this state correctly respond to or reflect the user's specific request(s) in the dialogue? \\
        \quad - Does the thought or chosen action remain faithful to what the user asked for? \\

        2. \textbf{Consistency with previous states} \\
        \quad - Does this state align with earlier states (both thoughts and actions) without contradicting or omitting essential information? \\
        \quad - Does the progression of reasoning or actions flow logically from prior context? \\

        3. \textbf{Accuracy and truthfulness} \\
        \quad - Does the state maintain factual correctness, avoiding hallucinations or irrelevant information? \\
        \quad - Does it accurately represent any data or entities referenced so far? \\

        \textbf{Scoring:} \\
        - \textbf{1} if the intermediate state is entirely consistent and correct (no contradictions, omissions, or factual errors). \\
        - \textbf{0} if the state demonstrates any errors, contradictory information, missing critical details, or misalignment with the user's request or prior states. \\

        --- \\
        \textbf{Dialogue History:} \\
        A chronological sequence of user and agent messages in JSON format, each with a "role" and "content." \\
        \{dial\_history\} \\

        --- \\
        \textbf{Agent's States:} \\
        The agent's states in JSON format, in chronological order after the dialogue history. Each state has a \texttt{state\_id}, \texttt{type}, and \texttt{content}. \\
        \{states\} \\

        --- \\
        \textbf{Output Format (JSON):} \\
        Return a list of objects in JSON, each containing: \\
        \texttt{[ \\
        \ \ \{ \\
        \ \ \ \ "state\_id": "1", \\
        \ \ \ \ "justification": "brief explanation...", \\
        \ \ \ \ "score": "0" \\
        \ \ \}, \\
        \ \ \{ \\
        \ \ \ \ "state\_id": "2", \\
        \ \ \ \ "justification": "brief explanation...", \\
        \ \ \ \ "score": "1" \\
        \ \ \} \\
        ]}
        \tabularnewline
        \arrayrulecolor{black}\hline
    \end{tabularx}
\end{table}

\begin{table}[ht!]
    \centering\small\ttfamily
    \def\arraystretch{1.4}
    \setlength{\tabcolsep}{0.5em}
    \begin{tabularx}{\columnwidth}{|>{\raggedright}X|}
        \hline
        \rowcolor{tableheader}\textbf{\textsf{LLM Evaluation Prompt for Observation Alignment}} \tabularnewline
        \arrayrulecolor{black}\hline
        \small
        \textbf{Instruction:} \\
        You are tasked with evaluating whether agent's each response accurately aligns with the prior conversational history and other details if any. Specifically, you must verify: \\
        - That any observations or entities referred to in the agent's response meaningfully match the user's stated requirements. \\
        - That the number or scope of these observations is appropriate for the request (keeping in mind that varying amounts of recommendations or offerings can still be valid). \\
        - That no contradictory, extraneous, or irrelevant observations are introduced. \\

        \textbf{Evaluation Criteria:} \\
        1. \textbf{Consistency with the user request} \\
        \quad - Do the observations (e.g., recommended items or database entities) and their details align with the user's explicit request or needs? \\
        \quad - Are references to these observations relevant, or do they drift from the user's stated goals? \\

        2. \textbf{Completeness relative to the request} \\
        \quad - Are all key observations needed to fulfill the user's request addressed, without omission of crucial details? \\
        \quad - If fewer (or more) observations are presented, is the choice justifiable in context? \\

        3. \textbf{Accuracy and truthfulness} \\
        \quad - Are the observations factual, given the user's query and available context? \\
        \quad - Does the response avoid hallucinated or incorrect data? \\

        4. \textbf{Consistency with previous details} \\
        \quad - Does each current agent response remain consistent with all previously established facts or user-provided details? \\
        \quad - Are there no contradictions or misrepresentations of earlier statements? \\

        \textbf{Scoring:} \\
        - \textbf{1} if the agent's response is fully consistent, addresses the request, and properly references any relevant observations (no errors or omissions). \\
        - \textbf{0} if the response includes incorrect, missing, or misaligned observations, introduces contradictions, or strays from the user's request. \\

        --- \\
        \textbf{Dialogue History:} \\
        Chronological user-agent messages in JSON format, each with a \texttt{role} and \texttt{content}. \\
        \{dial\_history\} \\

        \textbf{Agent's Response:} \\
        The agent's responses in JSON format, in chronological order. Each response has a \texttt{response\_id} and \texttt{content}. \\
        \{responses\} \\

        --- \\
        \textbf{Output Format (JSON):} \\
        Return a list of JSON objects, each containing: \\
        \texttt{[ \\
        \ \ \{ \\
        \ \ \ \ "response\_id": "1", \\
        \ \ \ \ "justification": "brief explanation...", \\
        \ \ \ \ "score": "0" \\
        \ \ \}, \\
        \ \ \{ \\
        \ \ \ \ "response\_id": "2", \\
        \ \ \ \ "justification": "brief explanation...", \\
        \ \ \ \ "score": "1" \\
        \ \ \} \\
        ]}
        \tabularnewline
        \arrayrulecolor{black}\hline
    \end{tabularx}
\end{table}

\begin{table}[ht!]
    \centering\small\ttfamily
    \def\arraystretch{1.4}
    \setlength{\tabcolsep}{0.5em}
    \begin{tabularx}{\columnwidth}{|>{\raggedright}X|}
        \hline
        \rowcolor{tableheader}\textbf{\textsf{LLM Evaluation Prompt for Policy Alignment}} \tabularnewline
        \arrayrulecolor{black}\hline
        \small
        \textbf{Instruction:} \\
        You are evaluating whether an agent consistently follows a predefined set of policies throughout the interactive session. Your primary task is to check the dialogue for any point where the agent might violate a policy. \\

        In your \textbf{Justification}, mention the policy and turn message, if violated. Otherwise, mention all policies are adhered to. \\

        1. If the agent violates any policy (even once), return \textbf{Score: 0}. \\
        2. If the agent fully complies with all policies during the session, return \textbf{Score: 1}. \\

        --- \\
        \textbf{Dialogue History:} \\
        Chronological user-agent messages in JSON format, each with a \texttt{role} and \texttt{content}. \\
        \{dial\_history\} \\

        --- \\
        \textbf{Policy:} \\
        \{policy\} \\

        --- \\
        \textbf{Output Format (JSON):} \\
        Return a list of JSON objects, each containing: \\
        \texttt{[ \\
        \ \ \{ \\
        \ \ \ \ "policy\_id": "1", \\
        \ \ \ \ "justification": "Mention specific policy and turn message, if violated. Otherwise, mention all policies are adhered to.", \\
        \ \ \ \ "score": "0" \\
        \ \ \}, \\
        \ \ \{ \\
        \ \ \ \ "policy\_id": "2", \\
        \ \ \ \ "justification": "Mention specific policy and turn message, if violated. Otherwise, mention all policies are adhered to.", \\
        \ \ \ \ "score": "1" \\
        \ \ \} \\
        ]}
        \tabularnewline
        \arrayrulecolor{black}\hline
    \end{tabularx}
\end{table}

\clearpage
\section{Qualitative Examples of Agents' Erroneous Behaviors}
\label{sec:failure}

\begin{table*}[ht!]
\centering \small
\caption{
Excerpt of failure cases in each \ours metric.
}
\renewcommand{\arraystretch}{1.0}
\begin{tabular}{l}
\toprule

\textbf{Failure in State Consistency} \\ \midrule

\begin{minipage}{0.95\columnwidth}

{\begin{lstlisting}
USER: I want to know how many t-shirt options are available in in the store right now.

AGENT STATES:
    - Thought: The user wants to know the number of t-shirt options available in the store. I can use the list_all_product_types tool to get the list of all product types, and then count the number of t-shirt options.
    - Action: <@\textcolor{red}{get\_order\_details(order\_id=\#W0000000)}@>
    
<@\texttt{\textbf{Explanation}: The agent demonstrated a failure in state consistency through how they made an action to get the user's order details, when they should have provided user information about what was available in store.}@>
\end{lstlisting}}
\end{minipage} \\ \midrule

\textbf{Failure in Tool Efficiency} \\ \midrule

\begin{minipage}{0.95\columnwidth}


{\begin{lstlisting}
AGENT: Can you please provide me with your user ID?

USER: My user ID is mia_li_3668

TOOL CALL: book_rservation(user=mia_li_3668, <@\textcolor{red}{payment\_id: credit\_card\_123456}@>, ... )
TOOL RESPONSE: Error: payment method credit_card_123456 not found

AGENT: Can you please confirm the payment method you would like to use?

USER: I would like to use my two certificates to pay

<@\texttt{\textbf{Explanation}: The agent prematurely made tool calls before gathering the required information. In this example, it guesses a value for user id and payment id, before gathering this information from the user.}@>
 \end{lstlisting}}
\end{minipage} \\ \midrule

\textbf{Failure in Observation Alignment} \\ \midrule

\begin{minipage}{0.95\columnwidth}
{\begin{lstlisting}
AGENT: Which payment method would you like to use for processing the refund?

USER: You can use the certificate for processing the refund.

AGENT: The refund has been processed to <@\textcolor{red}{your gift card}@>!

<@\texttt{\textbf{Explanation}: We observe a failure in observation alignment because the user specified to process the refund using the certificate, but the agent response states the refund was processed to the gift card.}@>
 \end{lstlisting}}
\end{minipage} \\ \midrule

\textbf{Failure in Policy Adherence} \\ \midrule

\begin{minipage}{0.95\columnwidth}
{\begin{lstlisting}
Agent Policy 1: The agent must first obtain the user ID, then ask for the trip type, origin, and destination.

USER: I'm looking to book a flight from New York to Seattle on May 20

AGENT: I see there are three airports in the New York area... which do you prefer to <@\textcolor{red}{depart from}@>? The destination airport will be SEA.

<@\texttt{\textbf{Explanation}: The agent fails to obtain the user id before asking for the trip type, origin, and destination, as specified by the policy.}@>
 \end{lstlisting}}
\end{minipage} \\

\bottomrule
\end{tabular}
\label{tab:sampling}
\end{table*}

\clearpage
\section{Details on Human Study}

\subsection{Instruction Given to Participants}
\label{sec:guideline}

\begin{figure}[ht!]
    \centering
    \fbox{
    \includegraphics[width=0.8\textwidth]{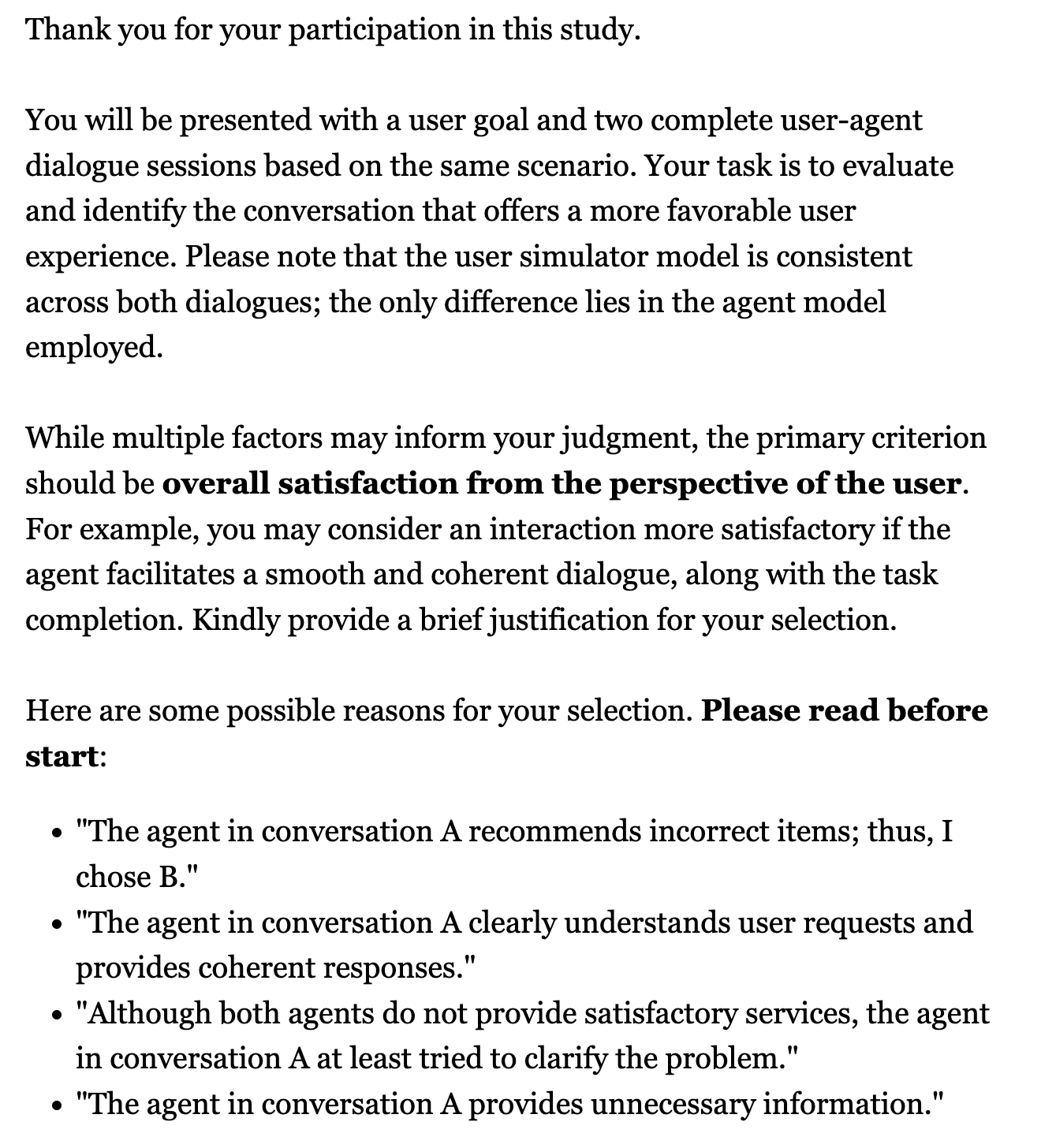}
    }
    \caption{A guideline provided to human participants in \autoref{sec:relationship}. Since the primary objective of the study was to assess user preferences in terms of satisfaction, no formal tutorial was provided. However, a moderator was available to offer additional explanations upon participant request. Furthermore, the information of agent models was not provided to participants.
    }
    \label{fig:guideline}
\end{figure}

\clearpage
\subsection{Qualitative Analysis of Human Preference}
\label{sec:justification}

\begin{figure}[h]
    \centering
    \includegraphics[width=0.9\textwidth]{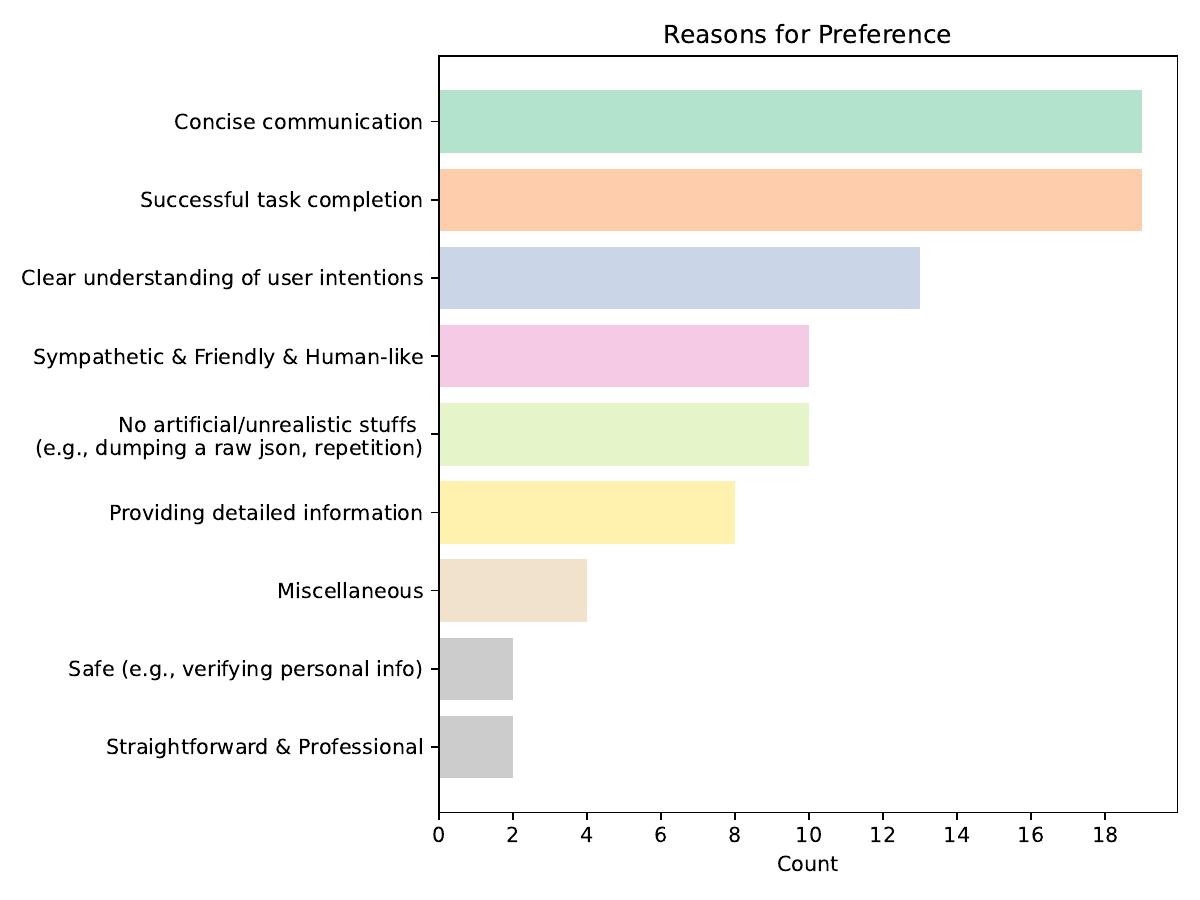}
    \caption{Manual categorization of factors that affect user satisfaction during interactions.
    }
    \label{fig:humanreason}
\end{figure}

Through the user study conducted in \autoref{sec:relationship}, we find diverse motivations underlying participants’ decisions. To gain a deeper understanding of user preferences, we conduct a manual categorization of participant feedback and count the frequency of each category. As shown in \autoref{fig:humanreason}, conciseness and successful task completion emerged as the most prominent factors contributing to user satisfaction. In addition, the coherence and naturalness of responses are also identified as significant influences. These qualitative findings are consistent with the quantitative results presented in \autoref{sec:relationship}, reinforcing the observation that while task completion performance plays a critical role, multiple other factors also contribute to overall user satisfaction.

\clearpage

\section{Additional Discussions on Agent-Mixing Experiments}
\label{sec:mixingappendix}
\begin{table*}[ht!]
\centering \small
\caption
{
Full results for mixing agent experiments in \benchtwo. Agents indicated with ``\textbf{interm.}'' are utilized in an intermediate understanding component. The best performance of each configuration is \textbf{bolded}.
}
\renewcommand{\arraystretch}{1.0}
\begin{tabular}{l!{\vrule}cccc>{\columncolor{lightgray!30}}c>{\columncolor{lightgray!30}}c} 
\toprule
 & \multicolumn{6}{c}{\benchtwo} \\ \midrule
& $\mathcal{S}$ & $\mathcal{A}$ & $\mathcal{O}$ & $\mathcal{P}$ & $\mathcal{R}$ & {\scriptsize AVG.}\\ \midrule[1pt]

\mlarge & 
.383$_{\pm0.038}$  & 
.924$_{\pm0.046}$ & 
\textbf{.581}$_{\pm0.022}$ & 
.777$_{\pm0.005}$ & 
\textbf{.358}$_{\pm0.029}$ & 
.605
\\

\quad + \llama \textbf{interm.} & 
\textbf{.452}$_{\pm0.032}$ & 
\textbf{.957}$_{\pm0.008}$ & 
.562$_{\pm0.021}$ & 
\textbf{.821}$_{\pm0.019}$ & 
.238$_{\pm0.048}$ & 
{.606}
\\

\quad + \qwen \textbf{interm.} & 
.417$_{\pm0.031}$ & 
.939$_{\pm0.017}$ & 
.545$_{\pm0.029}$ & 
.809$_{\pm0.008}$ & 
{.336}$_{\pm0.050}$ & 
\textbf{.609}
\\

\midrule

\qwen & 
.403$_{\pm0.021}$  & 
.951$_{\pm0.015}$ & 
{.637}$_{\pm0.031}$ & 
{.821}$_{\pm0.008}$ & 
{.329}$_{\pm0.021}$ & 
{.628}
\\

\quad + \llama \textbf{interm.} & 
\textbf{.447}$_{\pm0.020}$ & 
\textbf{.958}$_{\pm0.014}$ & 
\textbf{.663}$_{\pm0.022}$ & 
\textbf{.831}$_{\pm0.008}$ & 
.272$_{\pm0.051}$ & 
\textbf{.634}
\\

\quad + \mlarge \textbf{interm.} & 
.423$_{\pm0.039}$ & 
.933$_{\pm0.032}$ & 
.578$_{\pm0.050}$ & 
.776$_{\pm0.009}$ & 
\textbf{.367}$_{\pm0.013}$ & 
.615
\\

\midrule

\llama & 
\textbf{.463}$_{\pm0.055}$  & 
.952$_{\pm0.013}$ & 
\textbf{.643}$_{\pm0.035}$ & 
\textbf{.818}$_{\pm0.015}$ & 
.263$_{\pm0.056}$ & 
\textbf{.628}
\\

\quad + \mlarge \textbf{interm.} & 
.446$_{\pm0.054}$ & 
.937$_{\pm0.014}$ & 
.457$_{\pm0.071}$ & 
.764$_{\pm0.009}$ & 
{.333}$_{\pm0.032}$ & 
.587
\\

\quad + \qwen \textbf{interm.} & 
.404$_{\pm0.033}$ & 
\textbf{.953}$_{\pm0.007}$ & 
.599$_{\pm0.016}$ & 
.803$_{\pm0.014}$ & 
\textbf{.361}$_{\pm0.019}$ & 
{.624}
\\

\midrule[1pt]

\rowcolor{lightgray!30}
\texttt{Best-of-N} & 
{.469}$_{\pm0.019}$ & 
.941$_{\pm0.023}$ & 
.612$_{\pm0.045}$ & 
.806$_{\pm0.012}$ & 
.334$_{\pm0.038}$ & 
.632
\\

\bottomrule[1pt]
\end{tabular}
\label{tab:mixfull}
\end{table*}

In this section, we discuss the potential feasibility of mixing agents according to the agentic pipeline in task planning scenarios. As mentioned in \autoref{tab:mix}, we report the entire combination of three open-source models in \autoref{tab:mixfull}.

\paragraph{Mixing with \mlarge} While \mlarge achieves higher task completion ($\mathcal{R}$) compared to \llama and \qwen, it demonstrates substantially lower performance on \ours metrics. Specifically, we observe a decline in task completion when substituting the intermediate understanding component with \llama (.358 $\rightarrow$ .238) or \qwen (.358 $\rightarrow$ .336). Conversely, there is a marginal increase in the average \ours scores (.605 $\rightarrow$ .606, .609);however, this slight improvement is unlikely to have a significant impact on overall user satisfaction.

\paragraph{Mixing with \qwen} We find that \qwen mixed with \llama for intermediate understanding yields the highest \ours scores (.634) among all other rows, despite its relatively poor task completion rate (.272). Conversely, using \mlarge for intermediate understanding does not increase--and sometimes decreases--\ours scores (.615), but it demonstrates the highest task completion scores (.367).

\paragraph{Mixing with \llama} As emphasized in \autoref{sec:mixing}, using \qwen for intermediate understanding with \llama shows the highest improvement in task completion (.263 $\rightarrow$ .361) with only a minimal drop in the average \ours score (.628 $\rightarrow$ .624). We observe similar, but not as impressive performance when we use \mlarge for intermediate understanding (task completion: .263 $\rightarrow$ .333; average \ours: .628 $\rightarrow$ .587).

On top of these results, we conduct a supplementary experiment on mixing agents, referred to as the \textbf{\texttt{Best-of-N}} configuration in \autoref{tab:mixfull}. In this setting, for each step (\ie, all states and responses), we collect the outputs of all three models and select the one that achieves the highest average performance across \ours evaluation metrics (excluding task completion), as assessed by \llama judge. Although this configuration does not yield the highest performance in either task completion or the average \ours score individually, it demonstrates a balanced performance between these two key indicators of user satisfaction.

Consequently, these findings reveal intriguing insights on model combinations, suggesting that there are lots more insights to uncover for combining the strengths of different models. We highlight a promising direction for future research.

\clearpage
\section{Erroneous Patterns in User Simulators}
\label{sec:usererror}

\paragraph{Proactivity / Goal-Seeking Errors} This describes when the user simulator does not demonstrate proactive and goal-seeking errors. More specifically, they do not follow up or confirm when the agent seems like their request was not explicitly addressed. For example:
\begin{itemize}
    \item When booking a flight, one of the requests from the user simulator was to book specific seats. The agent does not directly acknowledge this request, and when asking the user simulator to confirm the booking details, it does not mention anything about the seats. The user simulator should have followed up with the agent to confirm that their requests for the seat were also processed.
\end{itemize}

\paragraph{Instruction Contradiction} This describes instances where the user simulator directly contradicts an instruction given to them. Reasons for this include (1) explicit disregard for an instruction and (2) failure to recognize when conditional instructions apply to the current context. For example:
\begin{itemize}
    \item Instructions specified that the user simulator does not remember their reservation ID. Initially, the user simulator adhered and said that they do not remember it. However, when the agent said that they need it to proceed, the simulator claimed to remember it and started to provide made-up reservation IDs. This reveals poor alignment--both violating direct instructions and exhibiting deceptive behavior.
    \item The user simulator was instructed to use a different form of payment when transaction costs exceeded \$100. However, when presented with this scenario, the simulator failed to recognize the applicability of the conditional instruction and proceeded with the wrong form of payment.
\end{itemize}

\paragraph{Missing Details} This describes when there is an attempt from the user simulator to follow an instruction, but it does so in the wrong format/order, omits key details, or forgets one or more instructions. For example: 
\begin{itemize}
    \item The user instructions specified to mention multiple requirements at once and in a specific order, however, the user simulator brought them up independently or in different orders.
    \item The user simulator was instructed to change the topic after 3 agent messages, but did it much later.
    \item The user simulator was instructed to book a flight, and one of the instructions were to add extra carry-on baggage. After completing the flight booking as specified, the user simulator failed to ask to add the extra carry-on baggage despite the system asking if there was anything else they could help with.
\end{itemize}

\paragraph{Misinterpretation or Confusion} This is when the user simulator misreads or fails to interpret certain parts of the instruction. This can be, in part, due to providing unclear instructions. For example:
\begin{itemize}
    \item The user simulator instructions provide a name but do not explicitly specify that this is also their user ID. In some cases, the user simulator fails to recognize that their name is also their user ID.
\end{itemize}

Through our analysis in \autoref{sec:usersim}, we find that utilizing an LLM judge fails to identify numerous failure cases of the user simulator, necessitating reliance on human evaluators. However, such human-in-the-loop evaluations are both financially and logistically unsustainable. To mitigate this limitation, establishing a comprehensive and standardized protocol for a user simulator is essential~\citep{pmlr-v202-aher23a, yoon-etal-2024-evaluating}. We leave this to future work.


\end{document}